\documentclass[10pt,twocolumn,letterpaper]{article}

\usepackage{cvpr}              

\definecolor{cvprblue}{rgb}{0.21,0.49,0.74}
\usepackage[pagebackref,breaklinks,colorlinks,allcolors=cvprblue]{hyperref}

\def\paperID{11719} 
\def\confName{CVPR}
\def\confYear{2025}

\usepackage{multirow} 
\usepackage{graphicx}
\usepackage{amsmath}
\usepackage{amssymb}
\usepackage{booktabs}
\usepackage[utf8]{inputenc} 
\usepackage[T1]{fontenc}    
\usepackage{amsfonts}       
\usepackage{nicefrac}       
\usepackage{microtype}      
\usepackage{algpseudocode}
\usepackage{times}
\usepackage{bbding}
\usepackage{paralist, tabularx}
\usepackage{caption}
\usepackage{pifont}
\usepackage{enumitem}

\usepackage{color, colortbl}
\usepackage{tabu}
\usepackage{makecell}
\usepackage{paralist}
\usepackage{threeparttable}

\usepackage{wrapfig}
\usepackage{subfloat}
\usepackage{floatrow}
\usepackage{algorithm}

\usepackage{xspace}
\usepackage{natbib}
\usepackage{appendix}
\usepackage{array,etoolbox}
\usepackage{tcolorbox}

\newcommand{\myparagraph}[1]{\vspace{0.3em}\noindent\textbf{#1}}

\definecolor{Gray}{gray}{0.3}

\definecolor{comm}{gray}{0.5}

\title{\model{}: Adaptive Tree-based Video Representation \\ for LLM Reasoning on Long Videos}

\author{%
Ziyang Wang\thanks{Equal contribution.} \quad 
Shoubin Yu$^*$ \quad 
Elias Stengel-Eskin$^*$ \quad \\
Jaehong Yoon \quad  
Feng Cheng \quad  
Gedas Bertasius \quad 
Mohit Bansal \\ 
UNC Chapel Hill \\
\\
{{ \tt \normalsize \href{https://videotree2024.github.io/}{\textcolor{magenta}{https://videotree2024.github.io/}} }}}

\definecolor{darkgreen}{rgb}{0,0.5,0}
\definecolor{azureblue}{rgb}{0,0.5,1}
\definecolor{darkgreen}{rgb}{1,0,0}
\definecolor{color1}{HTML}{006EB8}
\definecolor{color2}{HTML}{009B55}

\usepackage{pifont}

\usepackage[capitalize]{cleveref}
\crefname{section}{Sec.}{Secs.}
\Crefname{section}{Section}{Sections}
\Crefname{table}{Table}{Tables}
\crefname{table}{Tab.}{Tabs.}
\crefname{appendix}{Sec.}{Secs.}
\Crefname{appendix}{Section}{Sections}

\usepackage{color,colortbl}
\definecolor{darkgreen}{rgb}{0,0.5,0}

\definecolor{darkgreen}{rgb}{1,0,0}

\definecolor{azureblue}{rgb}{0,0.5,1}

\newcommand{\model}{\textsc{VideoTree}\xspace}

\begin{document}
\maketitle
\begin{abstract}
Long-form video understanding is complicated by the high redundancy of video data and the abundance of query-irrelevant information. 
To tackle these challenges, we propose \model{}, a training-free framework which builds a query-adaptive and hierarchical video representation for LLM reasoning over long-form videos. 
First, \model{} extracts query-relevant information from the input video through an iterative process, progressively refining the selection of keyframes based on their relevance to the query. 
Furthermore, \model{} leverages the inherent hierarchical structure of long video data, which is often overlooked by existing LLM-based methods. 
Specifically, we incorporate multi-granularity information into a tree-based representation, allowing \model{} to extract query-relevant details from long videos in a coarse-to-fine manner. 
This enables the model to effectively handle a wide range of video queries with varying levels of detail.
Finally, \model{} aggregates the hierarchical query-relevant information within the tree structure and feeds it into an LLM reasoning model to answer the query.
Our experiments show that our method improves both reasoning accuracy \emph{and} efficiency. 
Specifically, \model{} outperforms existing training-free approaches on EgoSchema and NExT-QA with less inference time, achieving $61.1\%$ and $75.6\%$ accuracy on the test set without additional video-specific training. 
Moreover, on the long split of Video-MME (average 44 minutes), \model{} achieves better performance than GPT-4V and many other MLLMs that were extensively trained on video data.
\end{abstract}

\section{Introduction}
\label{sec_intro}

\begin{figure}[t]
    \centering
    {\includegraphics[width=0.9\textwidth]{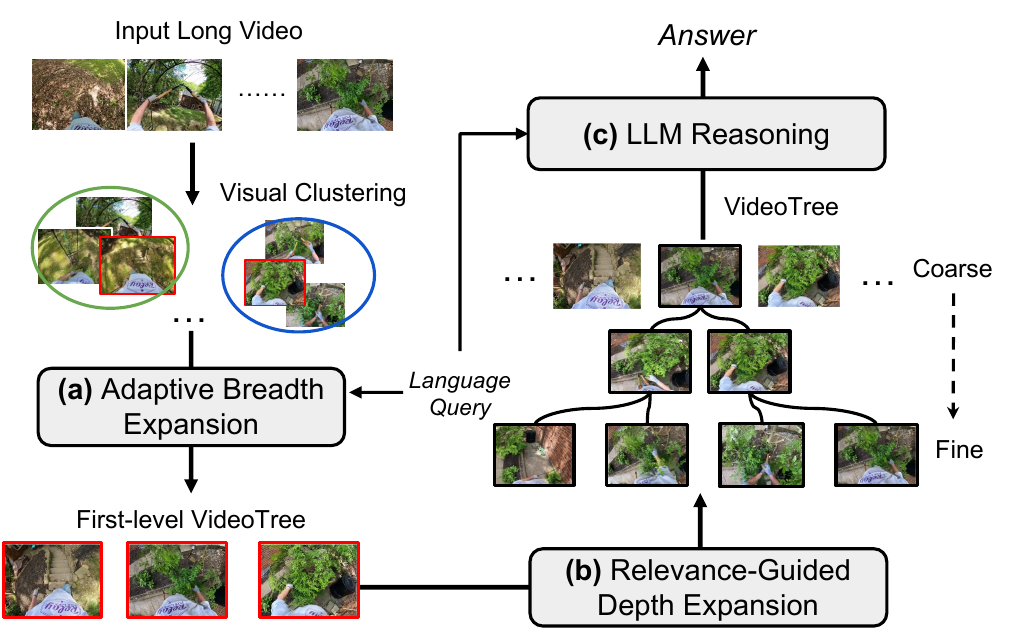}}
     \caption{Overview of \model{} for LLM reasoning on long videos. Given the long video input, we first apply adaptive breadth expansion to identify the first-level keyframes for \model{}. Next, we use relevance-guided depth expansion to explore the inherent hierarchical structure of the video, forming a tree-based representation. Finally, the coarse-to-fine information extracted by \model{} is fed into the LLM reasoner to answer the query.}
     \label{fig:teaser}
\end{figure}

With the surge in accessible long video content and the growing importance of applications such as long-form human behavior analysis and movie analysis, developing models capable of reasoning over and answering questions about long-form videos has become increasingly crucial. 
Recently, several approaches \citep{zhang2023simple, wang2024lifelongmemory, kahatapitiya2024language} have emerged that leverage the long-sequence reasoning capabilities of Large Language Models (LLMs) to tackle the challenge in long-form video understanding in a training-free manner.
Typically, these approaches leverage vision-language models (VLM) to caption densely sampled frames, thus representing the video in text format. 
This text representation is then subsequently fed into an LLM, which reasons over the video and responds to the provided query. 
Although this strategy has demonstrated great potentials on long-form video understanding benchmarks, it still faces two major limitations:

\noindent\textbf{1)  Informational Overload:} 
Long videos inherently contain high levels of information redundancy, and current approaches \citep{zhang2023simple, chung2023long} lack a principled method to effectively address this challenge. 
A deluge of redundant and irrelevant information can overwhelm the LLM, leading to mistakes in long-form video reasoning and reduced efficiency.

\noindent\textbf{2) Inability to Capture the Coarse-to-Fine Video Structure:} Existing approaches \citep{zhang2023simple, wang2024videoagent} often simplify video content into a list of captions without any structure, failing to account for the hierarchical nature of video information. 
Especially in long videos, some regions are information-dense -- requiring fine-grained temporal understanding -- while others are irrelevant to the query, or information-sparse.
Because of this, existing approaches not only suffer from overload problems, as mentioned above, but also omit key information from the captions, leading to missed details. 

These limitations underscore the pressing need for a new long-form video understanding method. 
To this end, we introduce \textbf{\model{}}, a training-free framework for long-form video understanding. \model{} dynamically extracts query-relevant keyframes from the video input in a coarse-to-fine manner and organizes them within a tree structure, with child nodes representing more fine-grained information. 
\model{} is \emph{adaptive}, meaning that our method allocates more frames to relevant video regions and fewer frames to irrelevant ones based on the given query.  
\model{} is also \emph{hierarchical}. 
Unlike existing approaches \citep{zhang2023simple, wang2024videoagent}, which treat video as a list of frames, we explore the inherent structure within the video data (e.g., events, scenes) to extract fine-grained information relevant to the query.

\model{} relies on three crucial steps: \textbf{adaptive breadth expansion} (\cref{fig:teaser}a), \textbf{relevance-guided depth expansion} (\cref{fig:teaser}b), and \textbf{LLM-based reasoning} (\cref{fig:teaser}c). 
To address redundancy in long videos, \model{} first leverages an adaptive breadth expansion module to extract query-relevant information, forming the initial level of representation. We utilize an iterative process of visual clustering, keyframe captioning, and relevance scoring until sufficient query-relevant information is gathered.
Compared to existing approaches \citep{kahatapitiya2024language, zhang2023simple} that rely on dense frame captions, \model{} selects only sparse keyframes for captioning, which significantly improves inference efficiency and helps avoid irrelevant information that could interfere with accurate video reasoning.
To capture more fine-grained information, we introduce a relevance-guided depth expansion step that adds finer, query-specific details in a hierarchical structure, forming a tree-based representation. 
Finally, we generate video descriptions from the structured representation using a captioner and provide them, along with the query, to the LLM for long video reasoning.

We demonstrate the effectiveness and efficiency of \model{} by evaluating it on two mainstream long video question answering (LVQA) datasets, EgoSchema \citep{mangalam2024egoschema}and NExT-QA \citep{xiao2021next}. 
Compared existing training-free approaches, \model{} achieves $2.1\%$ and $4.3\%$ improvements on EgoSchema(subset) and NExT-QA validation set with less inference time or LLM calls. 
To further validate \model{} effectiveness on very long videos, we test our method on the long split of the recent Video-MME benchmark~\citep{fu2024video} and \model{} achieves better performance than the strong proprietary GPT-4V model. 
Our ablation studies show that \model{} outperforms the the same category methods (VideoAgent~\citep{wang2024videoagent} and LLoVi~\citep{zhang2023simple}) under all number of captions and observes better efficiency-effectiveness trade-off. 
We further provide addition results on open-source LLM, where \model{} shows strong generalization ability across different language backbone models and achieves $4.8\%$ improvements against the LangRepo approach \citep{kahatapitiya2024language}.
\section{Related Work}
\paragraph{Structural Video Representation.}
Video understanding~\citep{lin2019tsm,wang2023masked,li2024videomamba, lin2023mmvid,xu2023retrieval,lin2023videollava,ma2023vistallama,wang2023gpt4video, wang2023chatvideo,wu2022memvit,ren2024timechat,song2024moviechat,liu2022unsupervisedtemporalvideogrounding,wu2024freevaofflinemllmtrainingfree} has shown impressive advancement in both views of comprehension and efficiency. Recently, several video-language methods \citep{ashutosh2023hiervl,li2020hero, islam2024video,zhang2018cross, Zala_2023_CVPR,qing2022learning, sanders2024tvtrees,yang2024doraemongpt,xiao2022hierarchical, lu2022lgdnlanguageguideddenoisingnetwork, xu2024slowfastllavastrongtrainingfreebaseline, wu2025longvituinstructiontuninglongform} have further introduced a structured understanding of video frames to allow compact and efficient recognition of scene contexts. For example, HierVL~\citep{ashutosh2023hiervl} proposes a \emph{bottom-up} hierarchical video-language embedding that capture video representations across short and long time periods. 
VideoReCap \citep{islam2024video} introduces a progressive video captioning approach that generates short clip-level captions and summarizes them into longer segments. 
These methods process long videos by progressively building high-level knowledge from local temporal information, i.e. in a bottom-up fashion that first captures all low-level details and then aggregates. 
This results in significant computational and time overhead. 
In contrast, inspired by the existing coarse-to-fine video understanding works~\citep{wu2019liteevalcoarsetofineframeworkresource, Wang_2023_ICCV}, \model{} proposes a novel top-down approach with a tree structure, enabling efficient and effective long video understanding by dynamically extracting query-relevant keyframes for LLM reasoning.

\paragraph{Video Understanding with LLMs.} 
Inspired by the powerful reasoning capabilities of LLMs, recent works have explored using LLMs to address complex video-related tasks. Since LLMs primarily process text, various methods~\citep{munasinghe2023pg,lin2023video, korbar2024textconditioned,weng2024longvlm, Maaz2023VideoChatGPT, zhang2023videollama, tan2024koala, chen2023video,li2024videochat,jin2024chatunivi,he2024malmm, li2024llms,wang2024lstp,li2024mvbench,yu2024crema，zhang2023movqabenchmarkversatilequestionanswering} have been developed to efficiently train multimodal projectors to connect the visual encoder and LLMs or leverage caption-centric information. 
Past works~\citep{wang2022longshort,kahatapitiya2024language,fan2024videoagent,wang2024videoagent, suris2023vipergpt, choudhury2023zero,wang2023vamos,ko2023large,wang2024lifelongmemory} has investigated training-free combinations of captioners and LLMs for video understanding.
Specifically, LLoVi~\citep{zhang2023simple} proposes a simple language-guided video understanding method. 
First, it extracts short-term video descriptions with a captioning model, and then an LLM summarizes these dense captions and responds to the given prompt. VideoAgent~\citep{wang2024videoagent} introduces a multi-round frame search strategy using an LLM agent. 
Unlike existing approaches, we propose a novel method to extract the key information from videos in an adaptive and coarse-to-fine manner with the agent, improving both efficiency and performance on long video understanding tasks. Moreover, \model{} improves interpretability by highlighting key visual clues for LLM reasoning via its human-readable tree structure.

\section{\model{} Method}

We present \model{}, a framework for constructing a query-adaptive, hierarchical video representation for efficient LLM reasoning over long videos. 
As illustrated in \cref{fig:framework_draft}, the \model{} framework consists of three main steps: adaptive breadth expansion, relevance-guided depth expansion, and LLM video reasoning.
Given the highly redundant nature of long videos, \model{} first leverages an adaptive breadth expansion module to extract query-relevant information from the video, forming the initial level of representation (\cref{sec: Adaptive Breadth Expansion}). 
To capture finer-grained details, we propose a relevance-guided depth expansion module that progressively adds finer-grained, query-specific details to in a hierarchical manner, forming a tree-based representation (\cref{sec: Relevance-Guided Depth Expansion}). 
Finally, we extract the video description from the constructed tree representation by using a captioner to caption selected frames. We feed it, along with the query, into the LLM for long video reasoning (\cref{sec: LLM video reasoning}).

\subsection{Adaptive Breadth Expansion}
\label{sec: Adaptive Breadth Expansion}
\begin{figure*}[t]
    \centering
    {\includegraphics[width=0.9\textwidth]{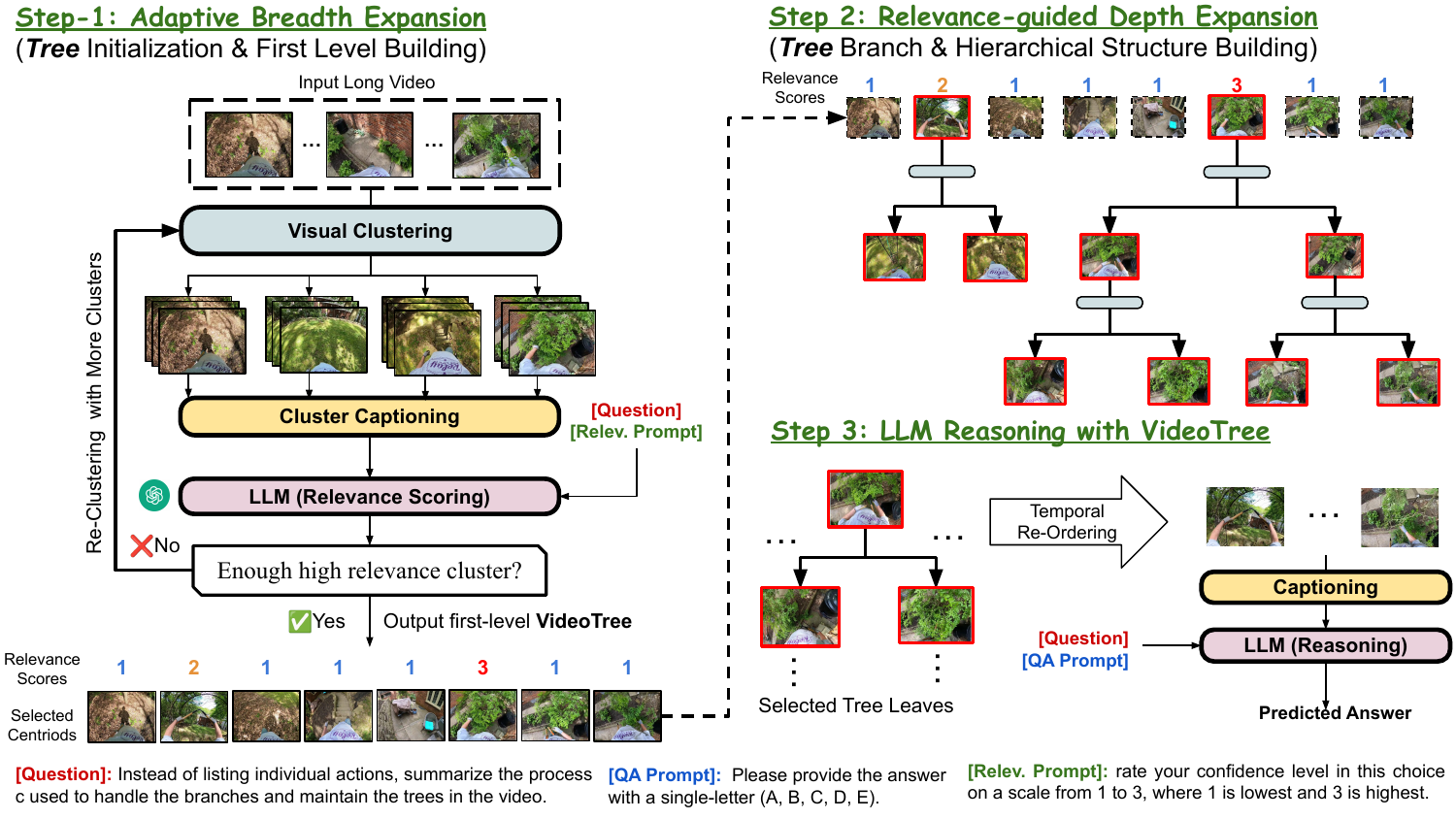}}
\caption{A detailed view of \model{}. To construct the tree structure, we begin with \emph{Adaptive Breadth Expansion} (Step 1), which dynamically extracts query-relevant key information, considering both video and question inputs. Then, starting from the highly relevant root nodes, we explore deeper into the tree branches with \emph{Relevance-guided Depth Expansion} (Step 2), re-clustering at each level to capture finer visual cues. Finally, we gather the selected nodes (keyframes), caption them, and arrange them in temporal order for \emph{LLM reasoning} (Step 3).}
    \vspace{-2mm}
    \label{fig:framework_draft}
\end{figure*}

Video data is often highly redundant, and long videos can contain substantial amounts of irrelevant information relative to the given video query. 
Addressing this redundancy and filtering out irrelevant content is crucial for efficient and effective long video understanding.
Existing approaches \citep{yu2024self, wang2024vila} select a fixed number of keyframes as the key information. However, as discussed in \cref{sec_intro}, this fixed keyframe selection is sub-optimal for a general long video-language understanding framework, since the information density varies across videos—some contain numerous scene changes, while others remain largely static.
To address this, we propose an adaptive breadth expansion module that constructs the first level of the tree representation by dynamically identifying keyframes that are relevant to the given query.
Specifically, as shown in the left of \cref{fig:framework_draft} (Step 1), given the video and a query about it, we build the first level of the tree by iterating three operations: \textbf{visual clustering}, \textbf{cluster captioning,} and \textbf{relevance scoring}. 
These operations first group similar frames together, then generate captions for each cluster, and use the LLM to determine how relevant each cluster is to the query. \model{} iterate these operations until getting enough query-relevant information from long videos in an \emph{adaptive} manner. 
In the following paragraphs, we provide a detailed motivation and introduction for each operation.

\paragraph{Visual Clustering.} 
To reduce the redundancy, we first propose a visual clustering operation that groups the video frames based on semantic similarity, allowing the model to focus on representative frames from each cluster while discarding repetitive or irrelevant content.
Specifically, given a video sequence $V=(F_1, F_2...,F_n)$, where $F_i$ is the frame at the time step $i$ and $n$ is the length of the video, we extract visual features for each frame with the pre-trained visual encoder~\citep{sun2024eva} $E$, such that $f_i = E(F_i)$, where $f_i\in\mathbb{R}^{d}$ is the visual features extracted by the frame $F_i$.
These features serve as a compact representation of each frame's visual content, capturing diverse semantics of each frame, such as scenes and objects. 
We then use K-Means clustering~\citep{macqueen1967some} to group frame features into $k$ distinct clusters based on their similarity, which we denote as: 
\begin{equation}
\scalebox{0.95}{$(C_1,\dots,C_k),\,(c_1,\dots,c_k)=\text{K-Means}((f_1,\dots,f_n),k)$}
\end{equation}

where, $C_i$ is the $i$th cluster that groups multiple frames, $c_i$ is the centroid vector for the $i$th cluster and $k$ is the number of clusters. 
This clustering process reduces the redundancy within the video by converting the input from $n$ frames into $k$ clusters of similar frames (where $n \gg k$), effectively summarizing the video into $k$ keyframes (cluster center frame) that capture the essential semantics.

\paragraph{Cluster Captioning.}
To better extract the key semantics from each cluster, we leverage a captioner to convert the keyframe information (a single frame or short clip around the keyframe) from each cluster to a textual description.
Specifically, for the cluster $C_i$, we find the keyframe $F_i$ that is closest to the centroid vector $c_i$ and consider it as the keyframe of the $i$th cluster. 
We then feed the extracted keyframe (or the key clip) into the VLM-based captioner $Cap(\cdot)$ \citep{zhao2023learning, liu2024llavanext} and obtain a text caption $t_{i} = Cap(F_i)$ for each cluster. These text captions serve as detailed descriptions of the key semantics from the corresponding clusters.

\paragraph{Relevance Scoring.}
To encourage the model to extract query-relevant information, after obtaining the cluster captions $t$, we leverage the reasoning capability of the LLM to assess whether the extracted information are sufficient for answering the given query.
To this end, we first feed all cluster captions $\{t_i \:\:\forall i \in [1, \ldots, k]\}$ from the last operation and the video query $Q$ into the LLM and output a set of relevance scores $\{r_i \:\:\forall i \in [1,\ldots, k]\}$ for each cluster, where $r_i$ is the relevance of the $i_{th}$ cluster.  
Specifically, to obtain each $r_i$, we prompt the LLM with the captions and the query, asking it to assign a relevance score to each caption, with three levels: 1 (not relevant), 2 (somewhat relevant), and 3 (highly relevant).
See \cref{tab:relevance_score_prompt} for all detailed prompts. 

Then, we adaptively extract the query-relevant information within the video by iterating the clustering, captioning, and relevance scoring operation. 
Specifically, given the list of relevance scores for each cluster, we set a threshold of the number of highly relevant clusters $rele\_num\_thresh$ to decide the stop of the adaptive process. 
We also set a maximum value for the number of clusters ($max\_breadth$) to avoid infinite loops. 
If the number of highly relevant clusters is below the requirement, that indicates the information extracted from the current cluster assignment is insufficient for the LLM to answer the video query. 
In that case, we increase the number of clusters $k$ by double the original number and repeat the clustering, captioning, and relevance scoring operations. 
If the number of high-relevance clusters meets the threshold $rele\_num\_thresh$ or the number of clusters reaches $max\_breadth$, we append the extracted clusters with their keyframes to the tree's first layer and continue to the next step (\cref{algo:our_method}, lines 2-11 for details).

\subsection{Relevance-Guided Depth Expansion}
\label{sec: Relevance-Guided Depth Expansion}

After obtaining the first-level clusters and their keyframes, \model{} captures high-level query-relevant information from the video input. However, some video regions are information-dense and critical for answering the query, requiring a more detailed selection of keyframes. 

Existing approaches, such as SeViLA \citep{yu2024self} and VideoAgent \citep{wang2024videoagent}, typically treat the selected frames as an unstructured list, overlooking the potential internal structure within the video data.
To address this, as shown in Step 2 of \cref{fig:framework_draft}, we construct a hierarchical video representation on top of the clusters from the previous breadth expansion step, allowing us to efficiently extract query-relevant details by leveraging the semantic relationships within the video data.
Specifically, we expand the depth of the tree by sub-clustering the clusters with higher relevance scores from the first step.
The intuition is that for high-relevance clusters, the LLM requires more detailed, granular information, while for low-relevance clusters, more information could actually lead to irrelevant details being included and could thus overwhelm the LLM, leading to incorrect reasoning.

To build the hierarchical structure, we use the relevance of a top-level cluster to determine how many levels of more granular information will be extracted from it.
Since the relevance score $r$ falls into one of three levels, we handle each first-level cluster differently based on its assigned relevance level.
For "somewhat relevant" clusters, we re-cluster the first-level cluster into $w$ sub-clusters, where $w$ represents the tree's branch width, ensuring that more keyframes are allocated to these moderately relevant clusters. 
For "highly relevant" clusters, we re-cluster into a two-level tree with a branch width of $w$ using hierarchical clustering while keeping the 1st-level cluster information from the previous K-Means step.
This coarse-to-fine exploration strategy allows for the detailed extraction of relevant information, supporting comprehensive video analysis for complex queries.
We repeat this process for all first-level medium- and highly-relevant clusters and build the hierarchical structure of \model{} (lines 12-15 in \cref{algo:our_method}).
After the breadth and depth expansion steps, we obtain the tree-based video representation for LLM reasoning over the long video.

\subsection{LLM Video Reasoning}
\label{sec: LLM video reasoning}

Finally, in order to use the LLM's ability on video reasoning, we need to present the LLM with a text-based video description. 
To this end, we traverse the nodes of the tree starting at the roots and expanding to the leaves, extracting keyframes from the tree's clusters at all levels and passing them into the captioner to obtain keyframe (short clip) captions. 
We then sort these keyframe (short clip) captions in temporal order and concatenate them into a textual description of the video. 
Finally, we pass this description and the input query to the LLM and output the final answer (see line 16-18 in \cref{algo:our_method}). 
Our full prompt is in \cref{tab:llm_reasoning}.
\section{Experimental Setup}

\paragraph{Tasks \& Datasets.} We test \model{} on three diverse long-form video question-answering benchmarks: 
(1) \textbf{EgoSchema}~\citep{mangalam2024egoschema}, a long-range video question-answering benchmark consisting of 5K multiple choice question-answer pairs spanning 250 hours of video and covering a wide range of human activities. 
Our ablation studies are conducted on the official validation set of EgoSchema which contains 500 questions (referred to as the EgoSchema Subset). The videos are 180 seconds long on average.
(2) \textbf{NExT-QA} \citep{xiao2021next}, a video question-answering benchmark for causal and temporal reasoning. It contains 5440 videos with an average length of 44s and approximately 52K questions. NExT-QA contains 3 different question types: Temporal (Tem.), Causal (Cau.), and Descriptive (Des.). 
(3) \textbf{Video-MME} \citep{fu2024video} is a recent-proposed comprehensive evaluation benchmark for video analysis. We test \model{} on the ``long-term videos'' split of the dataset (long split), whose average video length is 44 minutes, ranging from 30-60 minutes.

\paragraph{Implementation Details.}
We adopt GPT-4\footnote{version 1106} \citep{openai2023gpt4} as our LLM for all the main results. We also provide the results with open-source LLM (\cref{sec:analysis}) and other proprietary LLMs (\cref{sec:appendix_abalation}). Following VideoAgent \citep{wang2024videoagent}, we leverage EVA-CLIP-8B \citep{sun2024eva} as our visual encoder and also provide experimental analysis with smaller visual encoder in \cref{sec:analysis}.  
Following VideoAgent \citep{wang2024videoagent}, we leverage CogAgent \citep{hong2023cogagent} as the captioner for NExT-QA benchmark and use LaViLa~\citep{zhao2023learning} as our captioner for the EgoSchema benchmark due to its ego-centric video pretraining (we also show results in \cref{tab:captioner} using a unified captioner (LLaVA1.6-7B \citep{liu2024llavanext}) for all benchmarks).
For Video-MME, we directly use the default unified LLaVA1.6-7B captioner. 
We preprocess videos by simply sampling the original frames at 1FPS for EgoSchema and NExT-QA benchmark and 0.125 FPS for Video-MME. The best-performing average number of captions for EgoSchema subset, Next-QA and Video-MME is $62.4$, $12.6$ and $128$, respectively. We ablate our hyper-parameter choices in \cref{sec:appendix_abalation}.

\paragraph{Evaluation Metrics.}
We evaluate \model{} on all datasets under the multiple-choice QA setting. We utilize standard accuracy metrics for all experiments.

\section{Results}

\subsection{Comparison with Existing Approaches}

\begin{table*}[ht]
\RawFloats
\centering
\setlength{\tabcolsep}{4pt}

\begin{minipage}{0.65\textwidth}
\centering
\small
\begin{tabular}{lccccccccc}
\toprule
\multirow{2}{*}{\textbf{Model}} & \multicolumn{1}{c}{\multirow{2}{*}{\textbf{(M)LLM}}} & \multicolumn{2}{c}{\textbf{EgoSchema}}                                     &  & \multicolumn{4}{c}{\textbf{NExT-QA}} \\ \cmidrule(lr){3-4} \cmidrule(lr){6-9}
                                & \multicolumn{1}{c}{}                             & \multicolumn{1}{c}{\textbf{Sub.}} & \multicolumn{1}{c}{\textbf{Full}} &  & \multicolumn{1}{c}{\textbf{Tem.}} & \multicolumn{1}{c}{\textbf{Cau.}} & \multicolumn{1}{c}{\textbf{Des.}} & \textbf{Avg.}\\
\midrule

\rowcolor{gray!20} \multicolumn{9}{c}{{\textit{Based on Open-source Captioners and LLMs}}}   \\
MVU \citep{ranasinghe2024understanding} &Mistral-13B  & 60.3 & 37.6 & & 55.4 &48.1 &64.1 &55.2 \\
LangRepo \citep{kahatapitiya2024language} & Mixtral-8×7B & \textcolor{lightgray}{66.2\footnotemark[1]} & \textcolor{lightgray}{41.2} & & 51.4  & 64.4  &69.1 &60.9 \\
Video-LLaVA+INTP \citep{shang2024interpolatingvideollmslongersequencelmms} & Vicuna-7B v1.5 & - & 38.6 & & 58.6 &61.9 & 72.2 & 62.7 \\ 

\midrule
\rowcolor{gray!20} \multicolumn{9}{c}{{\textit{Based on Proprietary MLLMs}}}   \\
IG-VLM~\citep{kim2024image} & GPT-4V &  59.8&- & &  63.6&69.8&74.7&68.6 \\

\textcolor{lightgray}{LVNet~\citep{park2024framesusefulefficientstrategieslongform}~\footnotemark[2]}   & \textcolor{lightgray}{GPT-4o}  & \textcolor{lightgray}{68.2} & \textcolor{lightgray}{61.1} & & \textcolor{lightgray}{65.5} & \textcolor{lightgray}{75.0} & \textcolor{lightgray}{81.5} & \textcolor{lightgray}{72.9}\\

\midrule

\rowcolor{gray!20} \multicolumn{9}{c}{{\textit{Based on Open-source Captioners and Proprietary LLMs}}}   \\
ProViQ~\citep{choudhury2023zero} & GPT-3.5&  57.1&- & &  -&-&-&64.6\\
LLoVi~\citep{zhang2023simple} & GPT-3.5 &  57.6&50.3 & &  -&-&-&- \\
MoReVQA~\citep{min2024morevqa} & PaLM-2 &  -&51.7 & &  64.6&70.2&-&69.2\\
Vamos~\citep{wang2023vamos} & GPT-4 & 51.2 & 48.3  & &  -&-&-&- \\
LLoVi~\citep{zhang2023simple} & GPT-4 & 61.2 &- & &  61.0&69.5&75.6&67.7 \\
VideoAgent~\citep{wang2024videoagent} &GPT-4 &  60.2&54.1 & &  64.5 & 72.7 & 81.1 & 71.3\\
VideoAgent~\citep{fan2024videoagent} & GPT-4&  62.8&60.2  & &  -&-&-&-  \\
LifelongMemory~\citep{wang2024lifelongmemory}~\footnotemark[3] & GPT-4 & 64.1 &58.6 & &  -&-&-&- \\

\midrule
\model{} (Ours) & GPT-4 & \textbf{66.2} & \textbf{61.1} & & \textbf{70.6} & \textbf{76.5} &  \textbf{83.9} & \textbf{75.6} \\ 
\bottomrule
\label{tab:main_results}
\end{tabular}
\vspace{-1em}
\caption{Comparison with other training-free methods on EgoSchema and NExT-QA. \model{} outperforms the existing approaches on all evaluation metrics. }
\end{minipage}
\hfill
\begin{minipage}{0.3\textwidth}
\centering
\small
\begin{tabular}{lc}
    \toprule
    \textbf{Method} & \textbf{Acc} \\
    \midrule
    \rowcolor{gray!20} \textit{Proprietary MLLM} & \\
    GPT-4V & 53.5 \\
    GPT-4o & 65.3 \\
    Gemini 1.5 Pro & \textbf{67.4} \\                
    \midrule
    \rowcolor{gray!20} \textit{Open-Source MLLM} &  \\
    LongVA & 46.2 \\
    VITA & 48.6 \\
    InternVL2-34B & 52.6 \\
    VILA-1.5-40B & 53.8 \\
    Oryx-1.5-34B & 59.3 \\
    LLaVA-NeXT-Video-72B & 61.5	 \\
    Qwen2-VL-72B & \textbf{62.2} \\
    \midrule
    \rowcolor{gray!20} \multicolumn{2}{l}{\textit{Training-free Approach}} \\

    LLoVi & 48.8 \\
    \model{} (Ours) & \textbf{54.2} \\
    \bottomrule
\end{tabular}
\label{tab:videomme}
\caption{Video-MME long split results. \model{} outperforms the strong proprietary GPT-4V model and many other specially-trained open-souce video MLLMs (e.g. InternVL2-34B, VILA-1.5-40B) despite being training-free.}
\end{minipage}
\end{table*}

\paragraph{Comparison with training-free methods.} \cref{tab:main_results} shows a comparison of the existing training-free works and \model{} on EgoSchema and NExT-QA benchmarks. 
We compare our methods with three types of systems: those using all open-source LLMs \citep{ranasinghe2024understanding, kahatapitiya2024language, shang2024interpolatingvideollmslongersequencelmms}, those with proprietary MLLMs~\citep{kim2024image,park2024framesusefulefficientstrategieslongform}, and the most similar class to ours, which consists of methods with open-source captioners and proprietary LLMs \citep{choudhury2023zero, zhang2023simple, min2024morevqa, wang2023vamos, fan2024videoagent, wang2024videoagent, wang2024lifelongmemory}. Specifically, compared with the methods that leverage the same VLM (captioner) and LLM \citep{zhang2023simple, wang2024videoagent, wang2024lifelongmemory}, \model{} significantly outperforms these methods on both EgoSchema and NExT-QA benchmarks. Comparing with VideoAgent~\citep{fan2024videoagent} which also uses video-specific models (Video-LLaVA~\citep{lin2023videollava}, ViCLIP from InternVid \citep{wang2024internvidlargescalevideotextdataset}) which were trained on extensive video data, 
\model{} still performs better on EgoSchema. 
Moreover, comparing with the methods that utilize strong multimodal LLMs, \model{} significantly outperforms IG-VLM \citep{kim2024image} (based on GPT-4V\citep{2023GPT4VisionSC}) on both EgoSchema and NExT-QA benchmarks and obtains comparable results on the EgoSchema full test set compared to the recent LVNet \citep{park2024framesusefulefficientstrategieslongform} (which uses the more powerful GPT-4o for both captioner and LLM) while outperforming LVNet on NExT-QA benchmarks. 
Additionally, we observe a significant gap between \model{} and the open-source LLM-based approaches, highlighting the need of strong LLM reasoning module in our method. 
For the sake of making a fair comparison, 
we also show \model{}'s ability using open-source LLM in \cref{tab:open_source}, where we obtain an $4.8\%$ improvement on the EgoSchema subset. 
These results showcase the effectiveness of \model{} compared with existing training-free methods. 
Moreover, \model{} is also more efficient: we show analyses measuring the number of captions in \cref{fig:caption_num_abla} and inference time in \cref{tab:eff_eff_ana}, where \model{} is more efficient than relevant baselines.

\footnotetext[1]{We de-emphasize the EgoSchema results of LangRepo since it predicts the answers via a log-likelihood classifier rather than generation, making it different from all other methods (including \model{}).
We provide a comparison using the same classifier and LLM in \cref{tab:open_source} and show $4.8\%$ improvements under same settings.}

\paragraph{Evaluating on Very Long Videos.} To further highlight the strength of our approach on longer videos, we incldue results on Video-MME \citep{fu2024video}'s long split, which contains a diverse set of very long videos (up to 1 hour, with an average of 44 minutes). 
We compare our training-free method with three types of models, including proprietary MLLMs \citep{2023GPT4VisionSC, gpt4o, Reid2024Gemini1U} and open-source MLLM \citep{zhang2024longva, fu2024vitaopensourceinteractiveomni,chen2023internvl, chen2024far, liu2024oryx, wang2024vila, zhang2024llavanext-video, Qwen2VL}, both of which are trained on large-scale video(image) data, and training-free baseline LLoVi \citep{zhang2023simple}.  
As shown in \cref{tab:videomme}, compared to the training-free baseline, LLoVi, \model{} achieves a substantial $5.4\%$ improvement on the long split of the Video-MME benchmark, demonstrating its effectiveness in understanding videos across long time-scales.
Compared to proprietary MLLMs, \model{} outperforms the strong GPT-4V \citep{2023GPT4VisionSC} model by $0.7\%$. However, there is still a gap between \model{} and powerful long-context proprietary MLLMs (GPT-4o \citep{gpt4o}, Gemini 1.5 Pro \citep{Reid2024Gemini1U}).
When comparing to open-source MLLMs that were extensively trained on video data, our training-free \model{} method outperforms a number of these strong MLLMs including ViLA-1.5-40B \citep{lin2023vila} and Intern-VL2 \citep{chen2024far}.
\model{} achieves strong performance without additional training on long video data.

\footnotetext[2]{For fair comparison, we de-emphasize methods that use a much stronger MLLM (GPT-4o) as both the captioner and the LLM.}

\footnotetext[3]{Reproduced results, implementation details in \cref{sec:appendix_implement}}

\begin{table*}[]
\RawFloats
    \centering
    \begin{tabular}{lcccccc}
    \toprule
    \textbf{Method} & \textbf{Captions} & \textbf{Captioner (s)} & \textbf{Keyfr. (s)} & \textbf{QA (s)} & \textbf{Overall (s)} & \textbf{Acc.} \\
    \midrule 
    LLoVi-fast & 16 & 2.0 & 0 & 1.9 & \textbf{3.9} & 57.8 \\
    LLoVi-best & 180 & 22.4 & 0 & 2.4 & 24.8 & 61.2 \\ \midrule
    \model{}-fast & \textbf{13.6} & \textbf{1.6} & 4.4 & \textbf{1.8} & 7.8 & 63.6 \\
    \model{}-best & 62.4 & 7.8 & 10.2 & 2.1 & 20.1 & \textbf{66.2} \\
    \bottomrule
    \end{tabular}
    \caption{Efficiency-Effectiveness comparison between LLoVi and our approach. We benchmark the time cost of \model{} and LLoVi~\citep{zhang2023simple}, split into seconds spend in frame captioning, extracting keyframes, performing QA, and also report overall time. 
    Using only $33\%$ inference time, \model{}(fast) already achieves both better performance compared to LLoVi(best).}
    \label{tab:eff_eff_ana}
\end{table*}

\begin{table}[]
\RawFloats
\vspace{-1em}
    \centering
     \resizebox{0.85\columnwidth}{!}{ 
    \begin{tabular}{lccc}
    \toprule
    \textbf{Method}  & \textbf{\# Caption} & \textbf{ Acc.} & \textbf{Inf Time (s)} \\
    \midrule
\rowcolor{gray!20} \multicolumn{4}{l}{\textit{Based on Mistral-7B}} \\

    LLoVi & 180 & 50.8 & - \\
    LangRepo& 180 & 60.8 & 87.2 \\
    \model{} (ours) & 32 & \textbf{63.0} & \textbf{24.3} \\
    \midrule  
\rowcolor{gray!20} \multicolumn{4}{l}{\textit{Based on Mistral-8×7B (12B)}} \\
    LangRepo & 180 & 66.2 & 162.1 \\
    \model{} (ours) & 32 &  \textbf{71.0} & \textbf{50.3} \\
    \bottomrule
    \end{tabular}
    }
    \caption{Accuracy on the EgoSchema subset when using open-source LLM Reasoners and log-likelihood classifier. \model{} obtains better performance with less inference time on both 7B and 12B LLMs comparing to the LangRepo baseline \citep{kahatapitiya2024language}. }

    \label{tab:open_source}
\end{table}

\begin{table}[]
\RawFloats
    \centering
    \resizebox{0.8\columnwidth}{!}{ 
        \begin{tabular}{lc}
        \toprule
         \textbf{Module} & \textbf{ES Acc.} \\ \midrule
          \model{}  &  66.2 \\
          - Depth Expansion & 64.4 \\
          - Adaptive Breadth Expansion & 61.2 \\ 
        \bottomrule      
        \end{tabular}
    }
    \caption{Effect of different \model{} components. Both Adaptive Breadth Expansion and Depth Expansion modules contribute significantly to the effectiveness of \model{}.}

    \label{tab:ablation_module}
    \vspace{-3mm}
\end{table}

\subsection{Analysis}
\label{sec:analysis}
Below, we provide a detailed analysis of \model{} framework. All quantitative analyses are conducted on the validation subset of the EgoSchema dataset. First, we analyze the trade-off between efficiency and effectiveness, showing that our method has better efficiency \emph{and} performance across all settings compared to existing methods. We then provide an comprehensive ablation study for different design choice of \model{}. Finally, we visualize the tree from \model{} and show the clusters \model{} chooses to expand, qualitatively supporting its quantitative gains. 

\subsubsection{Efficiency-Effectiveness Analysis}

In \cref{tab:eff_eff_ana}, we show the efficiency-effectiveness trade-off of our approach compared to existing methods. 
Specifically, we compare \model{} with LLoVi \citep{zhang2023simple} using the same GPT-4 model as LLM (and same captioner). 
Comparing to the best model, LLoVi, \model{}-fast (which uses fewer frames by changing the hyper-parameters) achieves a $2.4\%$ improvement on the EgoSchema subset with only $33\%$ the time cost. 
Moreover, our best model obtains a $5.0\%$ improvement with less overall inference time compared to both LLoVi models. 
Profiling the inference time spent in different modules (including frame captioning, extracting keyframes/caption summarization, performing QA), we find that our hierarchical keyframe selection consumes a reasonable amount of time while significantly reducing the time cost in the captioning stage and boosting long video understanding performance. 
We also provide an ablation of average LLM calls and compared with VideoAgent~\citep{wang2024videoagent} in \cref{tab:ablation_llm_calls} showing that \model{} requires fewer LLM calls while having better performance. 
These results show that \model{} has better effectiveness and efficiency compared to the existing method.

\subsubsection{Ablation Study}

In this section, we conduct ablating different parts of \model on the EgoSchema subset. 
We ablate three features: Number of captions, applying open-source LLM and different \model{} components. 
We include more extensive ablations (including hyper-parameters and the design of captioner/LLM/vision encoder) in Appendix~\cref{sec:appendix_abalation}.

\begin{figure}[t]
    \centering
    \includegraphics[width=0.95\columnwidth]{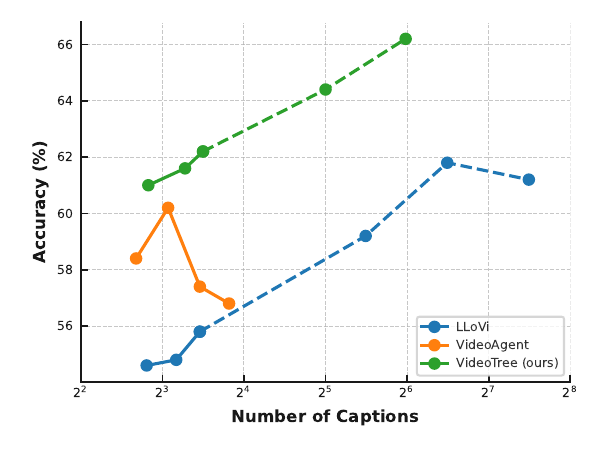}
    \caption{Ablating the number of captions. Given approximately the same number of frames, \model{} substantially outperforms LLoVi and VideoAgent. 
   Our hierarchical nature also allows it to generalize better to more frames and perform better overall.
    }
    \label{fig:caption_num_abla}
\end{figure}

\myparagraph{Number of Captions.\hspace{0.1em}} In \cref{fig:caption_num_abla}, we compare \model{} with existing methods under different caption settings. 
Under similar average frame caption settings (7, 9, 11), \model{} outperforms LLoVi~\citep{zhang2023simple} and VideoAgent~\citep{wang2024videoagent} by $6.5\%$ and $2.0\%$ on average accuracy across all three settings. 
Moreover, unlike the non-hierarchical VideoAgent baseline, which suffers from performance degradation after 11 frame captions (performing worse with $14$ frame captions), our method continues improving, generalizing to $62.4$ frame captions and achieving $6\%$ boost at its peak. 
It highlights the importance of \model{}'s hierarchical nature.

\definecolor{color2}{HTML}{009B55}
\begin{figure*}[ht]
    \centering
    {\includegraphics[width=0.93\textwidth]{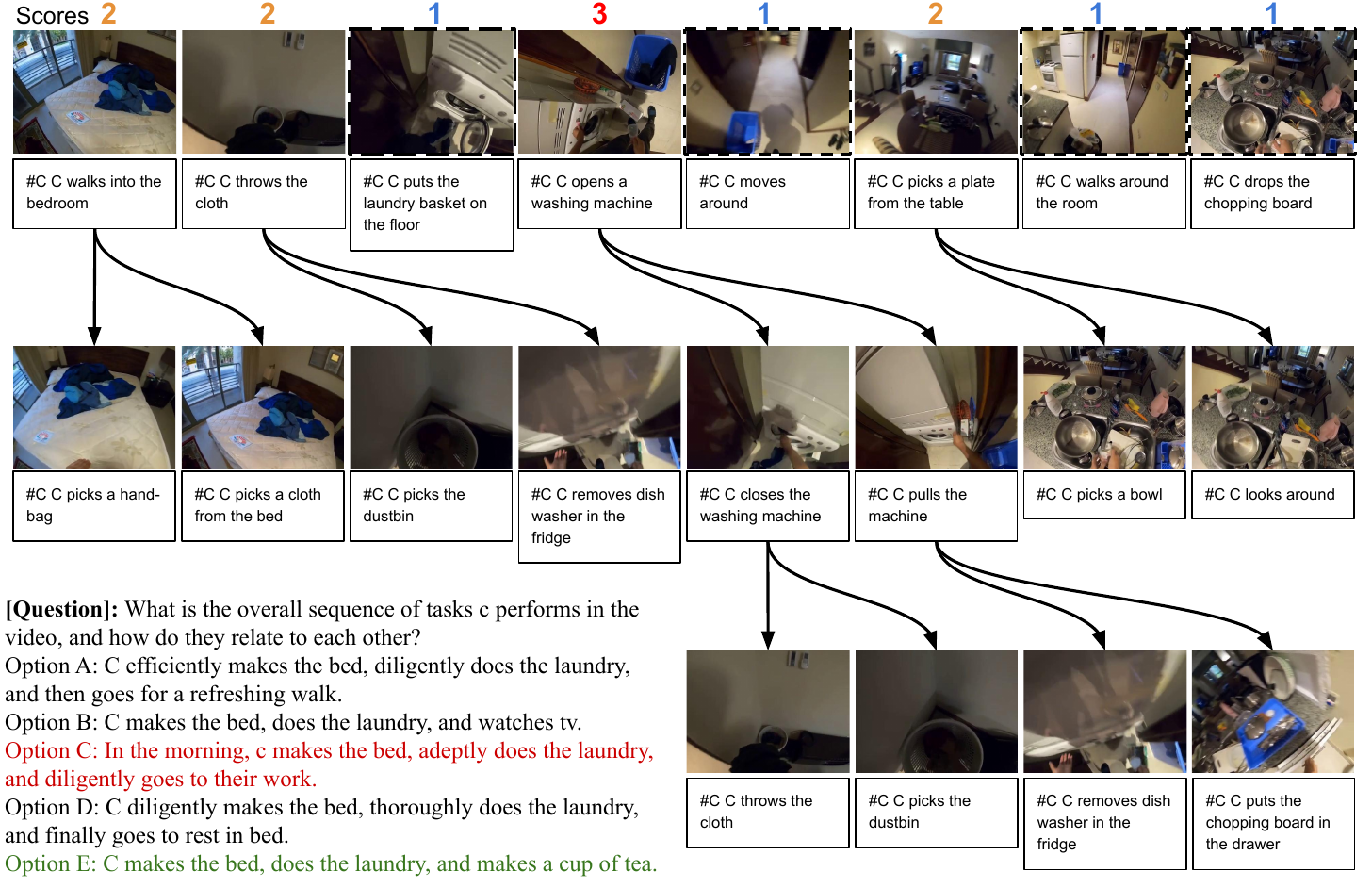}}
     \caption{Qualitative examples of \model{}.  \textcolor{red}{Red} options are answered wrongly with uniformly sampled 32 frames. 
  \textcolor{color2}{Green} options are answered correctly with \model{}. 
  Best viewed in color.}
    \label{fig:qualitative}
\end{figure*}

\myparagraph{Open-source LLM Reasoner.\hspace{0.1em}} To validate the effectiveness of \model{} with open-source LLM reasoners (rather than GPT4), 
in \cref{tab:open_source}, we report the performance of \model{} using 7B and 12B versions of the Mistral model~\citep{jiang2023mistral7b, jiang2024mixtralexperts} as the LLM reasoner.
We compare with LLoVi~\citep{zhang2023simple} and LangRepo~\citep{kahatapitiya2024language}. 
For a maximally fair comparison, we follow LangRepo's evaluation pipeline, using a log-likelihood classifier that scores all options and takes the highest-scoring one.  
\model{} substantially outperforms the baseline approaches on both 7B and 12B Mistral models while only requiring $20\%$ of the frame captions.
Specifically, compared to LangRepo, which uses complex textual summarization modules, \model{} achieves $2.2\%$ and $4.8\%$ better EgoSchema subset performance while using about $72.5\%$ and $69.0\%$ less inference time on Mistral 7B and 12B LLM, respectively. 
These results confirm that \model{}'s effectiveness and efficiency transfer to open-source models.

\myparagraph{\model{} Components.\hspace{0.1em}} In \cref{tab:ablation_module}, we report the effectiveness of the different components in \model{}. Specifically, removing the depth expansion module brings a $1.8\%$ drop in performance, showing the importance of the hierarchical design of \model{}. Removing the adaptive breadth expansion brings another $3.2\%$ decrease, verifying the effectiveness of the adaptive nature of \model{}.

\subsubsection{Qualitative Analysis}

In~\Cref{fig:qualitative}, we visualize qualitative results from \model{}. 
Specifically, we show the keyframes and their captions extracted by our adaptive tree representation given a video query. 
This example is drawn from EgoSchema, and shows the query format, which consists of a query and multiple-choice answers. 
With the proposed \model{} strategy, we split a complex multi-scene video (\eg \textit{cleaning house across rooms}) into several key scenes via visual clustering and determine the most query-relevant scene via the relevance score.
We then obtain more fine-grained visual cues by descending into each relevant cluster (Levels 2 and 3 in~\Cref{fig:qualitative}). 
For example \emph{``C opens a washing machine''} is deemed highly relevant to the question, which asks about the sequence of events. 
At the same time, frames like \emph{``C moves around''} are deemed irrelevant to the query and not expanded. 
In the end, \model{} shows a dynamic ability to select relevant segments and answer the given question correctly with only 50\% of the baseline's 32 input captions.
The LLoVi (fixed uniformly sampling) fails to correctly answer the question, sampling a large number of redundant and irrelevant frames. We also provide additional qualitative results in supplementary materials \cref{sec:append_qualitative}.
\section{Conclusion}
In this work, we proposed \model{}, an adaptive and hierarchical framework for LLM reasoning over long-form videos. 
\model{} adaptively extracts query-relevant keyframes from the video input in a coarse-to-fine manner and organizes them into a hierarchical representation, enabling the LLM to effectively handle complex queries. 
\model{} resulted in strong performance on three popular datasets (EgoSchema, NExT-QA, and Video-MME), while also improving efficiency by reducing the inference time and LLM calls. 
In our qualitative analysis, we showed that given a complex multi-scene video and its query, \model{} is capable of extracting key scenes and zooming into more detailed information that is highly related to the query. 
In the future, as more advanced captioners and stronger LLMs become available, the modular design of \model{} holds the potential for even greater performance and adaptability.

\section*{Acknowledgments}
We thank Ce Zhang, David Wan, and Jialu Li for their helpful discussions, and the reviewers for their feedback. This work was supported by DARPA ECOLE Program No. HR00112390060, NSF-AI Engage Institute DRL-2112635, DARPA Machine Commonsense (MCS) Grant N66001-19-2-4031, ARO Award W911NF2110220, ONR Grant N00014-23-1-2356, Sony Faculty Innovation award, Laboratory for Analytic Sciences via NC State University, and Accelerate Foundation Models Research program. The views contained in this article are those of the authors and not of the funding agency.

{
    \small
    \bibliographystyle{ieeenat_fullname}
    \bibliography{main}
}

\clearpage
\setcounter{page}{1}
\maketitlesupplementary

In this supplementary materials, we present the following: limitations (\cref{sec:appendix_limitation}), additional quantitative results (\cref{sec:appendix_quantitative}), 
additional ablation study for \model{} framework (\cref{sec:appendix_abalation}), 
the detailed algorithm for \model{} (\cref{sec:appendix_algo}),
additional implementation details (\cref{sec:appendix_implement}), additional qualitative analysis (\cref{sec:append_qualitative}).

\section{Limitations} 
\label{sec:appendix_limitation}
Like all LLM-based video-reasoning systems (including dense sampling) our method is limited by the ability of the captioner to accurately capture the contents of sampled frames. 
However, our method's modular nature means that as captioners improve, we can easily include them into the \model{} framework; similarly, we can use increasingly strong LLMs as the reasoning backbone of \model{}. 
While \model{} is training-free, it includes a small number of hyperparameters. 
In \cref{sec:appendix_abalation}, we ablate these hyperparameters, showing that \model{} outperforms the uniform-sampling baseline regardless of the choice of max depth and branch width. 
Thus, while better hyperparameters can benefit the method, even with sub-optimal settings \model{} outperforms the uniform baseline.

\section{Additional Quantitative Results}

\label{sec:appendix_quantitative}

\subsection{Comparison with advanced VideoLLMs on EgoSchema and NExT-QA.} In \cref{tab:compare_videollm}, we compare \model{} with advanced VideoLLMs \citep{wang2022internvideo, wang2024internvideo2, damonlpsg2024videollama2, wang2024tarsierrecipestrainingevaluating,shen2024longvuspatiotemporaladaptivecompression, li2024llavaonevisioneasyvisualtask} on EgoSchema and NExT-QA benchmarks. Without any video-specific training, \model{} gets comparable performance on EgoSchema fullset and slightly worse results on NExT-QA results, comparison with the methods was trained on large-scale video data and massive GPU hours.

\subsection{Additional evaluation benchmarks.} 
\paragraph{IntentQA Results. }
We report the IntentQA~\citep{li2023intentqa} results of \model{} and compare with existing methods. 
We first introduce IntentQA~\citep{li2023intentqa}, which contains 4,303 videos and 16K multiple-choice question-answer pairs focused on reasoning about people's intent in the video. We perform a zero-shot evaluation on the test set containing 2K questions. The videos are more than 44s in average length. We compare our methods with both training-free \citep{kahatapitiya2024language, zhang2023simple, yu2024self, kim2024image} and fine-tuned baseline \citep{wang2023vamos, xiao2022videographtransformervideo}. 

As shown in \cref{tab:intentqa}, our training-free \model{} approach achieves $66.9\%$ zero-shot accuracy on the test set, surpassing the existing training-free approaches LLoVi \citep{zhang2023simple} with $2.7\%$ improvements and even closing the gap with finetuned method Vamos \citep{wang2023vamos}. This result shows that \model{} improves performance in answering questions about intent, which is challenging since intent understanding \citep{li2023intentqa} requires the model to understand the various video contexts, including the immediate communicative context, the shared experience, and the commonsense. 

\begin{table}[ht!]
    \centering
    \caption{Comparison with advanced VideoLLMs on EgoSchema and NExT-QA benchmarks.}
    \begin{tabular}{lcc}
        \toprule
        \textbf{Method} & \textbf{ES} & \textbf{NExT-QA} \\
        \midrule
        InternVideo  & 32.1  & -\\ 
        Tarsier-34B  & 61.7 &  79.2	\\
        VideoChat2 & 60.2 	& 61.7\\
        VideoLLaMA 2 & 63.9 & - \\
        LLaVA-OV-72B  & 62.0 &  -	\\
        LongVU  & 67.6 &  -	\\

        \midrule
        \textit{Training-free Methods} \\
        \model{} (ours) & 61.1  & 75.6 \\
        \bottomrule
    \end{tabular}
    \label{tab:compare_videollm}
\end{table}

\begin{table}[ht!]
    \centering
    \caption{IntentQA Results}
    \begin{tabular}{lc}
        \toprule
        \textbf{Method} & \textbf{Accuracy} \\
        \midrule
        \textit{Fine-tuned Method} \\
        VGT & 51.3 \\
        Vamos & 68.5 \\
        \midrule
        \textit{Training-free Methods} \\
        LangRepo& 59.1 \\
        SeViLA & 60.9	\\
        LLoVi & 64.0 \\
        IG-VLM & 64.2	 \\
        \midrule
        \model{} (ours) & 66.9 \\
        \bottomrule
    \end{tabular}
    \label{tab:intentqa}
\end{table}

\paragraph{Video-MME Short and Medium Split Results.}
In \cref{tab:mme_short_mvlu}, we test \model{} on the short and medium splits of Video-MME benchmark as well. We apply the recent LLaVA-OV-7B model \citep{li2024llavaonevisioneasyvisualtask} as the frame captioner. 
Specifically, \model{} achieves $67.8\%$ and $59.9\%$ on Video-MME short and medium split, a more than $5.7\%$ and $6.7\%$ improvement compared to LLoVi \citep{zhang2023simple} and LongVA \citep{zhang2024longva}.

\begin{table}[]
\centering
\begin{tabular}{lcc}
\hline
\textbf{Model} & \textbf{V-MME Short/Med} & \textbf{MLVU-Avg} \\
\hline
LLoVi    &       62.1/53.2          &   55.1    \\
LongVA   &       61.4/50.9      &   56.3    \\
\hline
\model{} &    \textbf{67.8}/\textbf{59.9}  &  \textbf{60.4} \\
\hline
\end{tabular}
\caption{Results on Video-MME (short and medium splits) and MLVU with LlaVA-OV-7B as captioner.}
\label{tab:mme_short_mvlu}
\end{table}

\paragraph{MLVU Results.}
In \cref{tab:mme_short_mvlu}, we test \model{} on the MLVU validation set. We again use LLaVA-OV-7B \citep{li2024llavaonevisioneasyvisualtask} as the frame captioner. Results show that \model{} achieves a \textbf{$5.3\%$} and  \textbf{$4.1\%$} gain over LLoVi \citep{zhang2023simple} and LongVA \citep{zhang2024longva}, respectively.

\section{Additional Ablation Study} 
\label{sec:appendix_abalation}
In this section, we report additional ablation studies for \model{} framework. First, we ablate the LLM calls and vision encoder size for our method. Then, we show the effect of the different hyperparameter settings for \model{}. Finally, we analyze the effect of different VLM/LLM designs for \model{}. 

\paragraph{LLM Calls.} In \cref{tab:ablation_llm_calls}, we report the number of average LLM calls of \model{} and compare with VideoAgent \citep{wang2024videoagent}.\model{} achieves better results on much less LLM calls under similar caption numbers (only about 30\% LLM calls are needed). This is due to the adaptive and hierarchical structure of \model{} which could extracts more keyframes faster instead of searching one frame a time. This results highlight the advantage of the hierarchical nature of \model{} in both efficiency and effectiveness, comparing to the non-hierarchical approaches.

\begin{table}
\RawFloats
        \centering

\begin{tabular}{ccc}
\toprule
\textbf{Caption Number} & \textbf{Avg LLM Calls} & \textbf{ES Subset Acc} \\ \midrule
\rowcolor{gray!20} \multicolumn{3}{c}{\textit{VideoAgent}} \\

6.4 &   10.2 &  58.4 \\
8.4 &  10.2  &  60.2 \\
11.0 &  9.0   & 57.4   \\ \midrule
\rowcolor{gray!20} \multicolumn{3}{c}{\textit{\model{} (ours)}} \\

7.1 &  \textbf{2.3}  & \textbf{61.0}  \\
9.7 &  \textbf{2.5}   & \textbf{61.6}   \\ 
11.3 &   \textbf{2.8}  & \textbf{62.2}   \\ \bottomrule      
\end{tabular}
        \caption{The comparison of average LLM calls for \model{} and VideoAgent~\citep{wang2024videoagent} (estimated) under similar frame settings on EgoSchema subset. Results show that \model{} achieves better results on much less LLM calls.}
\label{tab:ablation_llm_calls}
\end{table}

\paragraph{Visual Encoder.} In \cref{tab:visual_encoders}, we study the effect of the visual encoder used in the visual clustering operation. 
We report the results of \model{} on three different scales of visual encoder: OpenCLIP-B, OpenCLIP-G \citep{ilharco_gabriel_2021_5143773} and EVA-CLIP-8B \citep{sun2024eva} and compare to VideoAgent \citep{wang2024videoagent}~\footnote{Note that VideoAgent only report results on OpenCLIP-ViT-G (1B) and EVA-CLIP-8B.}.
\model{} outperforms VideoAgent by an average of $6.9\%$ across both encoders. 
Comparing different visual encoders ranging from 88M to 8B parameters, we see only a marginal drop in performance for \model{} as the visual encoders decrease in size, indicating that our approach generalizes well to much smaller vision encoders (i.e. only a 0.2\% drop when going from 8B to 88M), making the model more efficient while maintaining strong performance. 
Additionally, we test the performance of a self-supervised vision encoder for \model{}. 
Specifically, we apply DINOv2-base \cite{oquab2024dinov2learningrobustvisual} to \model{} and get $64.2\%$ on EgoSchema subset, which is $1.8\%$ lower than the same size of CLIP model.

\begin{table}[htbp]
    \centering
    \caption{Testing different visual encoder design choices in \model{}. We also compare with VideoAgent\citep{wang2024videoagent} to show the effectiveness of our method. }
    \resizebox{\columnwidth}{!}{ 
        \begin{tabular}{lccc}
        \toprule
        \textbf{Visual Encoder} & \textbf{Params} & \textbf{Method} & \textbf{Acc.} \\
        \midrule
        \multirow{2}{*}{OpenCLIP-ViT-B} & \multirow{2}{*}{88M} & VideoAgent & -- \\
        & & \model{} & 66.0 \\
        \midrule
        \multirow{2}{*}{OpenCLIP-ViT-G} & \multirow{2}{*}{1B} & VideoAgent & 59.2 \\
        & & \model{} & \textbf{66.2} \\ 
        \midrule
        \multirow{2}{*}{EVA-CLIP-8B} & \multirow{2}{*}{8B} & VideoAgent & 59.4 \\
        & & \model{} & \textbf{66.2} \\
        \bottomrule
        \end{tabular}
    }
    \label{tab:visual_encoders}
\end{table}

\paragraph{Hyperparameter Analysis.}  

\begin{table}[ht!]
    \centering
    \RawFloats
        \centering
        \begin{tabular}{lcc}
        \toprule
        \textbf{Branch Width} & \textbf{ES Acc$\uparrow$} & \textbf{\#Frame$\downarrow$} \\ \midrule
        2          &  64.4   &43.5   \\
        3          &  65.0   &54.6   \\
        4          &   \textbf{66.2}  &62.4 \\
        5          &  64.2   &72.5  \\ \midrule
        \rowcolor{gray!20} Uniform Baseline &  61.2  & 180 \\ \bottomrule
        \end{tabular}
        \caption{The effect of different settings for branch width of \model{}. When the branch width is set to 4, \model{} achieves the best performance on the EgoSchema subset. Reducing the branch width makes the model more efficient while retaining performance, outperforming all existing approaches.}
        \label{tab:ablation_branch_width}

\end{table}

In \cref{tab:ablation_branch_width}, we study the effect of the branch width of the tree-based representation for the \model{}. 
The best performance is obtained when the branch width is set to $4$. 
As with depth, excessive branch width reduces the \model{} performance due to the information overwhelming to the LLM; however, even with the worst branch width setting, \model{} still outperforms the baseline.

In \cref{tab:ablation_max_breadth}, we study the effect of the max breadth of the adaptive tree-based representation. The results indicate that even with a smaller max tree breadth, \model{} achieves good performance while using much fewer frames. Increasing the breadth generally increases performance, with the best performance when the max breadth is set to $32$. 
However, having an excessive max breadth leads to worse results, suggesting that incorporating too much information in the adaptive tree-based representation limits the LLM reasoning ability.
This links back to the intuition of having an efficient representation for the LLM reasoning over long videos. 

In \cref{tab:ablation_threshold}, we study the effect of the threshold on the number of highly relevant clusters, which controls the iterative process of the adaptive breadth expansion process. 
The best performance is obtained when the branch threshold is set to $4$. Reducing the threshold improves the efficiency while retaining strong performance compared to the baseline results.

\begin{table}[ht!]
    \centering
    \begin{tabular}{lcc}
        \toprule
        \textbf{Max Breadth} & \textbf{ES Acc} & \textbf{\#Frame} \\ \midrule
        8          &   63.0  & 26.9  \\
        16         &   64.0  & 49.0  \\
        32         &   \textbf{66.2}  & 62.4  \\
        64         &   63.2  & 94.6 \\ \midrule
        \rowcolor{gray!20} Uniform Baseline &  61.2  & 180 \\ 
        \bottomrule
    \end{tabular}
    \caption{The effect of different settings for the max breadth of the first level of the tree. Results show that when the max breadth is set to 32, \model{} obtains the best performance. Reducing the max breadth improves efficiency while retaining performance.}
    \label{tab:ablation_max_breadth}
\end{table}

\begin{table}[ht!]
    \centering
    \begin{tabular}{lcc}
        \toprule
        \textbf{Threshold} & \textbf{ES Acc} & \textbf{\#Frame} \\ \midrule
        2          &   63.6 & 13.9  \\
        3         &   64.4  & 32.2	  \\
        4         &   \textbf{66.2}  & 62.4  \\
        5         &   64.8  & 79.2 \\ \midrule
        \rowcolor{gray!20} Uniform Baseline &  61.2  & 180 \\ 
        \bottomrule
    \end{tabular}
    \caption{The effect of different settings for the threshold on the number of highly relevant clusters. Results show that when the threshold is set to 4, \model{} obtains the best performance. Reducing the threshold improves efficiency while retaining performance.}
    \label{tab:ablation_threshold}
\end{table}

\paragraph{VLM Captioner.} In \cref{tab:captioner}, we compares the performance of the best captioner (LaViLA for EgoSchema and CogAgent for NExT-QA) with using a LLaVA-1.6-7B \citep{liu2024llavanext} captioner everywhere. We observe a comparable performance on NExT-QA compared with the best captioner, while still outperforming all other existing methods in \cref{tab:main_results}. We also observe a drop in performance on the EgoSchema subset while using LLaVA-1.6 captioner, this is likely due to a lack of egocentric data during LLaVA training, which is needed for strong performance on EgoSchema. In the future, we would like to see strong unified captioner that operate well across datasets; these would fit seamlessly into the \model{} framework, further boosting the performance.
Additionally, we test the performance of the question-prompted captioner by adding the video query into the prompt of the LLaVA-1.6 captioner. The results show that on EgoSchema subset, \model{} with a question-prompted captioner achieves $63.2\%$ accuracy, $3.0\%$ lower than the direct captioner.

\begin{table}[]
\RawFloats
    \centering
    \begin{tabular}{lccc}
    \toprule
    \textbf{Captioner} & \textbf{EgoSchema Sub} & \textbf{NExT-QA} \\
    \midrule
    LLaVA-1.6-7B & 59.2 & 73.6 \\
    Best Model& \textbf{66.2} & \textbf{75.6}  \\
    \bottomrule
    \end{tabular}
    \caption{Comparing accuracy with \model{} using the same captioner throughout (LLaVA1.6-7B) and best captioner for each benchmarks. }
    \label{tab:captioner}
\end{table}

\paragraph{LLM Reasoner.} We ablate the design choice of captioner and LLM for the \model{} framework in \cref{tab:ablation_llm}. With a better LLM, \model{} can achieve better performance on long video understanding tasks, indicating the potential \model{} to improve as its modules become more advanced. Notably, our GPT-3.5 variant substantially outperforms existing methods with the same LLM and standard QA prompts (VideoAgent \citep{wang2024videoagent} $48.8\%$, LLoVi \citep{zhang2023simple} $51.8\%$), achieving $57.6\%$ accuracy on EgoSchema subset.

\paragraph{Tree structure.} We first note that our ablation (\cref{tab:ablation_module}, \textbf{1.8\%} improvements) highlights the importance of the tree structure to keyframe \emph{selection}.
To test its utility as a \emph{representation} for the reasoner LLM, we conducted additional ablations, giving the reasoner a version of the tree's caption linearized in a top-down left-right traversal.
On EgoSchema subset, this new version scores $64.8\%$ while the temporal ordering one scores $66.2\%$; 
thus, while the tree is key to keyframe selection, the reasoner benefits from temporal order in video tasks.

\begin{table}
\RawFloats
        \centering
\begin{tabular}{lcc}
\toprule
\textbf{Method} & \textbf{LLM} & \textbf{ES Acc} \\ \midrule
LLoVi &  GPT-3.5  &  51.2 \\
VideoAgent &  GPT-3.5  &  48.8 \\
\model{} (Ours) & GPT-3.5    & \textbf{57.6}   \\ \midrule
LLoVi &  GPT-4  & 61.2  \\
VideoAgent & GPT-4    & 60.2   \\ 
\model{} (Ours) & GPT-4    & \textbf{66.2}   \\ \bottomrule      
\end{tabular}
\caption{The effect of different design choices of the LLM Reasoner for \model{}.}
\label{tab:ablation_llm}
\end{table}

\section{Detailed Algorithm} 
\label{sec:appendix_algo}
In \cref{algo:our_method}, we present the algorithm behind \model{}.

\begin{algorithm*}
\caption{\model{}}
\begin{algorithmic}[1]
\Require Video frames $V$, query $Q$, number of clusters $k$, threshold for the number of high-relevance cluster $rele\_num\_thresh$, maximum number of clusters allowed $max\_breadth$, branch width $w$, visual encoder $E$, LLM $F_{llm}$, captioner $F_{vlm}$, cluster information $C$, relevance score $R$, tree-based video representation $Tree$

\State $ k \gets k\_init$
\While{$k \leq max\_breadth$} \Comment{Adaptive breadth expansion}
    \State $C \gets \text{VisualClustering}(E, V, k)$
    \State $Cap \gets \text{ClusterCaptioning}(F_{vlm}, V, C)$
    \State $R \gets \text{RelevanceScoring}(F_{llm}, C, Q, Cap)$
    \If{$\text{count}(r \in R \mid r = high) \geq rele\_num\_thresh$}
    \State $Tree \gets Tree.append(C)$  \Comment{First level of \model{}}
    \State \textbf{break}
    \EndIf
    \State $k \gets k * 2$
\EndWhile

\For{$C_i \in C $} \Comment{Relevance-guided depth expansion}
    \State $ \hat{C_i} \gets \text{DepthExpansion}(E, C_i, R_i,w)$
    \State $Tree \gets Tree.append(\hat{C_i})$  \Comment{Adding hierarchy of \model{}}
\EndFor

\State $Cap \gets \text{GetCaptions}(F_{vlm}, V, Tree)$ \Comment{LLM Reasoning}
\State $pred\_answer \gets \text{LLMReasoning}(F_{llm},Cap, Q)$
\State \Return $pred\_answer$
\end{algorithmic}
\label{algo:our_method}
\end{algorithm*}

\section{Additional Implementation Details} 
\label{sec:appendix_implement}

\textbf{Additional \model{} Implementation Details.} 
For clustering, we use the {\tt{kmeans\_pytorch}} library. 
The hyper-parameter setting for $max\_breadth$, $max\_depth$, $branch\_width$ and $rele\_num\_thresh$ on the EgoSchema and Video-MME benchmark is $32$, $3$, $4$ and $4$ and for NExT-QA, we set the hyper-parameter as  $8$, $3$, $2$, and $3$. 
The initial $k$ depends on the average video length, and is set to $4$ for NExT-QA and $8$ for EgoSchema and VideoMME.

\textbf{Lifelong Memory Reproduce Details.} In \cref{tab:main_results}, we report the main results of LifelongMemory ~\citep{wang2024lifelongmemory} which is lower than the number than they reported in their paper. Here, we introduce our reproduce process in detail. For captions, since LifelongMemory authors do not provide the exact caption data/path, we directly utilize the same captioner from \model{} method and all other existing works (LaViLA) and extract the captions by 0.5FPS according to LifelongMemory paper. We then use their code to run the experiments on EgoSchema, however, the results are low and we observed a low success rate of the QA process (only about 80\% success samples). We then update their output format/process code, which boost performance by about 10\% and get the results in \cref{tab:main_results}, but still lower than their paper results. Thus, for fair comparison, we directly reported the reproduced results. 

\begin{figure*}[t]
    \centering
    {\includegraphics[width=\textwidth]{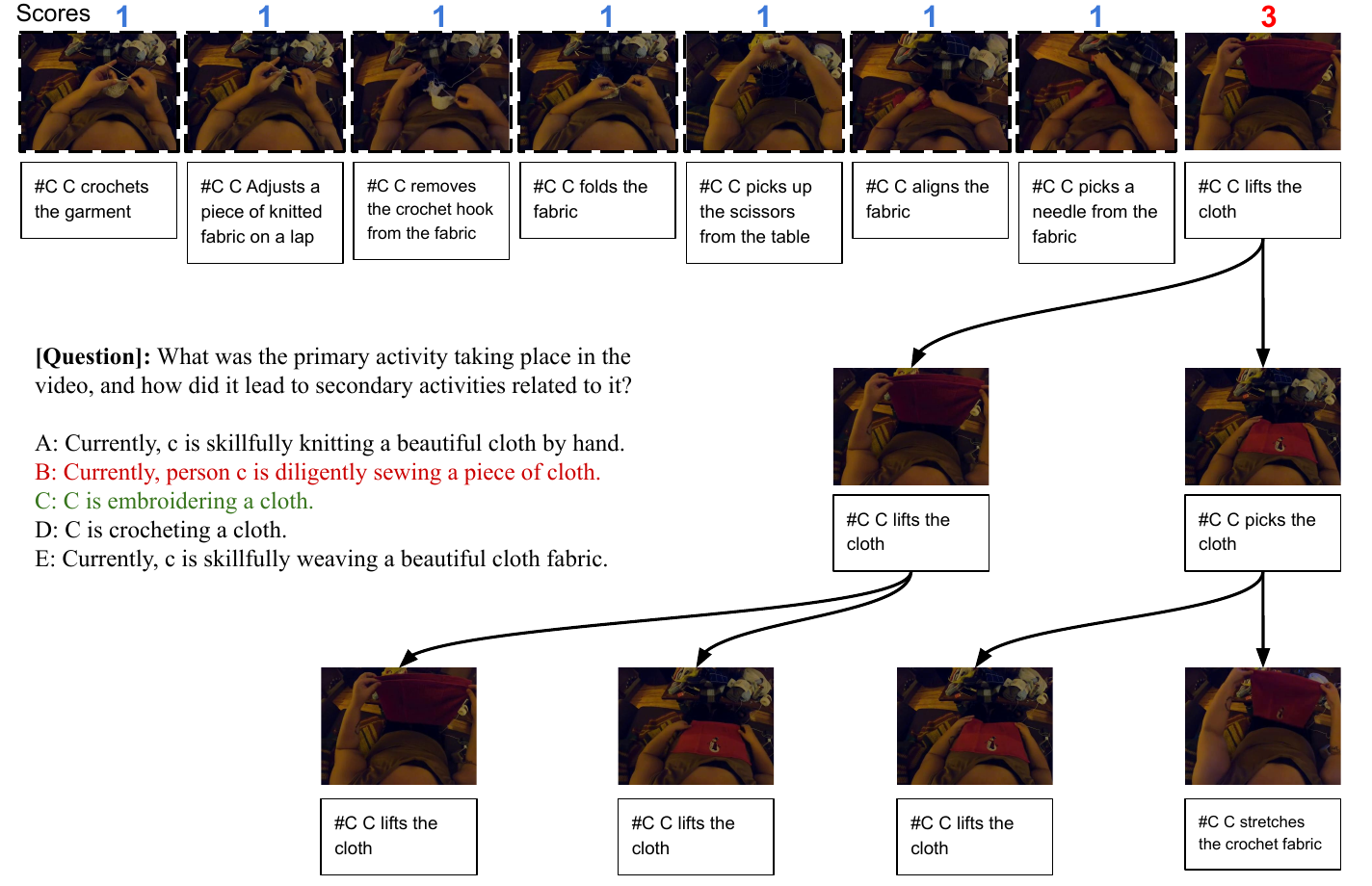}}
     \caption{Qualitative examples of \model{} keyframes and captions selection.  \textcolor{red}{Red} options are answered wrongly with uniformly sampled frames. 
  \textcolor{color2}{Green} options are answered correctly by \model{}. 
  Best viewed in color. }
    \label{fig:qualitative2}
\end{figure*}

\textbf{Prompt Details.}
We provide detailed prompts for the relevance scoring prompt in \cref{tab:relevance_score_prompt} and LLM reasoning prompt in \cref{tab:llm_reasoning} on the EgoSchema benchmark. 
\begin{table*}[t!]
\centering
\RawFloats
\begin{minipage}{1\columnwidth}\vspace{20mm}    
    \centering
    \caption{\textbf{\model{} with relevance scoring prompt on EgoSchema.}}
    \begin{tcolorbox} 
        \centering
        \hspace{-6mm}
        \begin{tabular}{p{0.99\columnwidth}}
        \hspace{1mm}
        \begin{minipage}{0.99\columnwidth}
        \textbf{User} \\
        You are presented with a textual description of a first view video clip, it consists of about \textcolor{blue}{\texttt{caption\_number}} frame captions sparsely sampled from the video (\#C means the first person view, and \#O indicates another). The ultimate goal is to answer a question related to this video, choosing the correct option out of five possible answers. \\
        It is crucial that you imagine the visual scene as vividly as possible to enhance the accuracy of your response. After selecting your answer, rate your confidence level in this choice on a scale from 1 to 100, where 1 indicates low confidence and 100 signifies high confidence. Please provide a concise one-sentence explanation for your chosen answer. If you are uncertain about the correct option, select the one that seems closest to being correct. Meanwhile, could you provide a relevance score for each frame caption to evaluate their relevance with the query-answering process. The score is between 1,2,3, where 1 indicates low relevance and 3 signifies high relevance. Please return the relevance score in the format of a list of \textcolor{blue}{\texttt{caption\_number}} scores. \\
        Examples:  \textcolor{blue}{\texttt{Examples}} \\
        Description:  \textcolor{blue}{\texttt{Captions}} \\
        Question: \textcolor{blue}{\texttt{Question}} \\
        Options: 
        A: \textcolor{blue}{\texttt{Option-A}}. B: \textcolor{blue}{\texttt{Option-B}}. C: \textcolor{blue}{\texttt{Option-C}}. D: \textcolor{blue}{\texttt{Option-D}}. E: \textcolor{blue}{\texttt{Option-E}}. \\
        The prediction, explanation, confidence and frame relevance are (please response in the format of 'prediction:, explanation:, confidence:, frame relevance:')\\
        \rule[0.25\baselineskip]{\textwidth}{1pt}
        \textbf{Assistant} \\
        {\textcolor{blue}{prediction}}, {\textcolor{blue}{explanation}}, {\textcolor{blue}{confidence}}, {\textcolor{blue}{frame relevance}}
        \end{minipage}
        \end{tabular}
    \end{tcolorbox}
    \vspace{-2mm}
    \label{tab:relevance_score_prompt}
\end{minipage}
\end{table*}

\begin{table*}[t!]
\centering
\begin{minipage}{1\columnwidth}\vspace{20mm}    
    \centering
    \RawFloats
    \caption{\textbf{\model{} with LLM reasoning prompt on EgoSchema.}}
    \begin{tcolorbox} 
    
        \centering
        \hspace{-6mm}
        \begin{tabular}{p{0.99\columnwidth}}
        \hspace{1mm}
        \begin{minipage}{0.99\columnwidth}
        \textbf{User} \\
        You are presented with a textual description of a first view video clip, it consists of frame captions sparsely sampled from the video (\#C means the first person view, and \#O indicates another). The ultimate goal is to answer a question related to this video, choosing the correct option out of five possible answers. \\
        It is crucial that you imagine the visual scene as vividly as possible to enhance the accuracy of your response. After selecting your answer, rate your confidence level in this choice on a scale from 1 to 100, where 1 indicates low confidence and 100 signifies high confidence. Please provide a concise one-sentence explanation for your chosen answer. If you are uncertain about the correct option, select the one that seems closest to being correct. \\
        Examples:  \textcolor{blue}{\texttt{Examples}} \\
        Description:  \textcolor{blue}{\texttt{Captions}} \\
        Question: \textcolor{blue}{\texttt{Question}} \\
        Options: 
        A: \textcolor{blue}{\texttt{Option-A}}. B: \textcolor{blue}{\texttt{Option-B}}. C: \textcolor{blue}{\texttt{Option-C}}. D: \textcolor{blue}{\texttt{Option-D}}. E: \textcolor{blue}{\texttt{Option-E}}. \\
        The prediction, explanation, and confidence is (please response in the format of 'prediction:, explanation: ,confidence:')\\
        \rule[0.25\baselineskip]{\textwidth}{1pt}
        \textbf{Assistant} \\
        {\textcolor{blue}{prediction}}, {\textcolor{blue}{explanation}}, {\textcolor{blue}{confidence}}
        \end{minipage}
        \end{tabular}
    \end{tcolorbox}
\label{tab:llm_reasoning}
\end{minipage}
\end{table*}

\textbf{Experiments Compute Resources.} All experiments are conducted on 4 (or less) NVIDIA-A6000 GPU and Azure Cloud APIs (for OpenAI models). The minimal GPU memory requirement is 24GB.

\section{Additional Qualitative Analysis} \label{sec:append_qualitative}

\textbf{Additional Visualization.}  
In \cref{fig:qualitative2} we show another visualization from \model{}. 
Here, \model{} localizes a single key activity (embroidering a cloth) taking place in the video and dynamically expands its constituent frames to answer the question correctly using a minimal number of frames. As shown in \cref{fig:qualitative,fig:qualitative2}, the chosen keyframe distribution depends on the query: general queries yield sparser keyframes that extend to distant parts of the video. Queries with detailed actions/objects yield more concentrated keyframes.

\begin{figure*}
    \centering
    \includegraphics[width=1\linewidth]{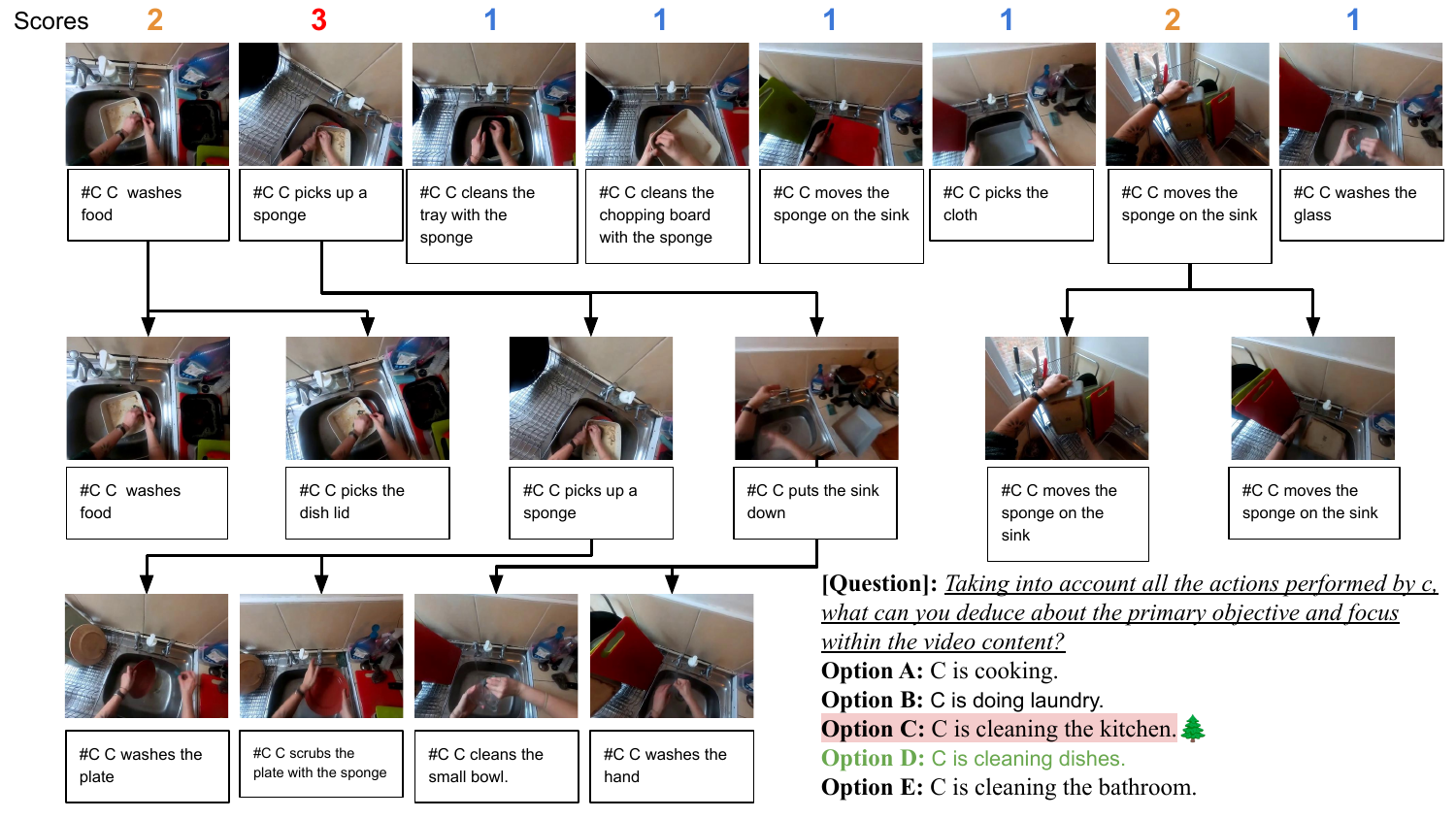}
    \caption{Failure case of \model{}.}
    \label{fig:failure}
\end{figure*}

\textbf{Failure Case.} We provide the qualitative visualization of a failure case in \cref{fig:failure}. Here, we find that the failure was due to the following factors:
A) The video had little scene change and multiple similar repeated actions (washing dishes).
B) As a result, when \model{} expands down to more fine-grained details, the captioners give detailed description that contain some hallucinations. These captions miss the correct higher-level keyword (dish) in the selected captions. With stronger captioners, this failure case could potentially be resolved.

\textbf{Human study.}
We conduct a human study on the accuracy of the keyframe scoring module to measure the quality of the LLM-based keyframe selection. 
Specifically, we ask a human annotator to judge the relevance of all 1st-level keyframes to the query, and we compare these decisions to the GPT4-based score used to evaluate this feature in \model{}.
On $345$ keyframes from $20$ EgoSchema videos, results show that the GPT4-based keyframe scoring achieves $90.7\%$ agreement with our human annotator, suggesting it captures human preferences well.

\end{document}